\documentclass[11pt,letterpaper]{llncs}
\usepackage{amsmath}
\usepackage{amssymb}
\usepackage{graphicx}
\usepackage{marvosym}
\usepackage{url}
\usepackage{fancyhdr}
\usepackage{multirow}

\usepackage{geometry}
\newcommand{\keywords}[1]{\par\addvspace\baselineskip
\noindent\keywordname\enspace\ignorespaces#1}

\begin{document}


\title{\LARGE{Analyzing Persuasive Strategies in Meme Texts: A Fusion of Language Models \\ with Paraphrase Enrichment}}


%
%
\author{\large{Kota Shamanth Ramanath Nayak \and Leila Kosseim}}
\institute{\large{Computational Linguistics at Concordia (CLaC) Laboratory \\
        Department of Computer Science and Software Engineering \\
        Concordia University, Montr\'eal, Qu\'ebec, Canada}}

%


%
%


\maketitle

\begin{abstract}
This paper describes our approach to hierarchical multi-label detection of persuasion techniques in meme texts. Our model, developed as a part of the recent SemEval task, is based on fine-tuning individual language models (BERT, XLM-RoBERTa, and mBERT) and leveraging a mean-based ensemble model in addition to dataset augmentation through paraphrase generation from ChatGPT. The scope of the study encompasses enhancing model performance through innovative training techniques and data augmentation strategies. The problem addressed is the effective identification and classification of multiple persuasive techniques in meme texts, a task complicated by the diversity and complexity of such content. The objective of the paper is to improve detection accuracy by refining model training methods and examining the impact of balanced versus unbalanced training datasets. Novelty in the results and discussion lies in the finding that training with paraphrases enhances model performance, yet a balanced training set proves more advantageous than a larger unbalanced one. Additionally, the analysis reveals the potential pitfalls of indiscriminate incorporation of paraphrases from diverse distributions, which can introduce substantial noise. Results with the SemEval 2024 data confirm these insights, demonstrating improved model efficacy with the proposed methods.

\end{abstract}


\section{Introduction}

The recent SemEval-2024 shared task 4~\cite{semeval2024task4} proposed three distinct subtasks dedicated to identifying persuasion techniques conveyed by memes. The primary aim was to unravel how memes, integral to disinformation campaigns, employ various techniques to shape user perspectives. Subtask~1 focused on the analysis of textual content alone and mandated the detection of 20 persuasion techniques structured hierarchically within the textual content of memes. On the other hand, subtasks~2 and~3 involved the analysis of multimodal memes that considered both textual and visual elements.\\This paper describes the approach we used for our participation to subtask~1 and further analysis with the dataset after the shared task. The task provided a training dataset in English, but in addition to being tested on English, the task also mandated the evaluation of our model's zero-shot performance in three surprise languages. The goal during the testing phase was not only to explore our model's ability to perform well in English but also generalize to other languages without explicit training.\\Inspired by successful approaches in multilabel text classification~\cite{jurkiewicz-etal-2020-applicaai,tian-etal-2021-mind}, our strategy involved fine-tuning three language models, i.e, BERT, XLM-RoBERTa, and mBERT, followed by ensemble modeling using mean aggregation. To enhance performance, we used data augmentation through paraphrasing and adjusted the classification thresholds for each persuasion technique based on class-wise metrics optimised using the validation set. During testing, a zero-shot approach based on machine translation was implemented to classify instances in the surprise languages. The official results of our system~\cite{semeval2024task4} demonstrated good performance advantages over the baseline in all languages except Arabic, where the increase in performance was not significant. Results also show that paraphrasing techniques enhances model performance but using a balanced training set is more beneficial than a larger unbalanced one. Furthermore, analysis highlights the potential downside of indiscriminately incorporating paraphrases from diverse distributions, as this can introduce notable noise into the system.\\This paper is organized as follows: Section~\ref{sec:pw} reviews relevant previous work in the field of multi-label classification. Section~\ref{sec:bg} provides an overview of the task and the data utilized. Section~\ref{sec:sys} presents an overview of our classification pipeline along with the techniques used for data augmentation; while Section~\ref{sec:exp} describes experimental details. Finally, the analysis of our model's results is presented in Section~\ref{sec:res}; while Section~\ref{sec:conc} provides conclusions and outlines future work.

\section{Previous Work}\label{sec:pw}

Several studies have explored multi-label classification of textual content. In 2019,~\cite{chalkidis-etal-2019-large} showed that Bi-GRUs with label-wise attention led to good performance, and the inclusion of domain-specific Word2vec and context-sensitive ELMo embeddings further boosted the performance on the EURLEX57K dataset that contained 57k English EU legislative documents.~\cite{lin-etal-2023-effective} introduced five innovative contrastive losses for multi-label text classification using the dataset from the SemEval~2018 Multi-label Emotion Classification (MEC) task~\cite{SemEval2018Task1} in English, Arabic, and Spanish that contained 8640 instances. All five contrastive learning methods notably enhanced the performance of the previous top-performing model, SpanEmO~\cite{alhuzali-ananiadou-2021-spanemo}, for the MEC task. Among these approaches, the Jaccard Similarity Probability Contrastive Loss demonstrated the highest effectiveness on the English dataset, achieving $F_{Macro}$ and $F_{Micro}$ scores of 57.68 and 71.01, respectively. 

In hierarchical multi-label classification (HMC), samples are assigned to one or more class labels within a structured hierarchy. Approaches to HMC can be divided into local and global methods. Local methods use multiple classifiers, often overlooking the overall structure of the hierarchy. For example, \cite{CERRI201439} trained a multi-layer perceptron incrementally for each hierarchy level, using predictions from one level as inputs for the next. In contrast, global methods employ a single model to address all classes and implement various strategies to capture the hierarchical relationships between labels. One such approach \cite{zhou-etal-2020-hierarchy} modeled the hierarchy as a directed graph and introduced hierarchy-aware structure encoders, using a bidirectional TreeLSTM and a hierarchy-GCN to extract and aggregate label structural information in an end-to-end fashion. \cite{Yu2022ConstrainedSG} redefined hierarchical text classification (HTC) as a sequence generation task and developed a sequence-to-tree (Seq2Tree) framework to model the hierarchical label structure. Additionally, they created a constrained decoding strategy with a dynamic vocabulary to ensure label consistency.

In the context of the SemEval~2020~Task~11~\cite{da-san-martino-etal-2020-semeval}, two subtasks were introduced addressing span identification of propagandistic textual fragments  and a multi-label technique classification (TC) of propagandistic fragments using  a corpus of 7k instances from the news domain. The subsequent SemEval~2021 Task~6~\cite{dimitrov-etal-2021-semeval} focused on the identification of propagandistic techniques from multimodal data including text and images from memes. The top-performing teams in 2020 and 2021, ApplicaAI~\cite{jurkiewicz-etal-2020-applicaai} and MinD~\cite{tian-etal-2021-mind} respectively, leveraged pre-trained language models and ensemble techniques to achieve top scores at the shared tasks. This year's shared task was built upon the 2021 task but included hierarchical metrics as well as a multilingual setting and a new training dataset of 7k instances in English. Inspired by the top models at \cite{da-san-martino-etal-2020-semeval} and \cite{dimitrov-etal-2021-semeval}, our methodology is also based on an ensemble of pre-trained language models but leverages paraphrase generation to further improve performance.
\section{Labelling Persuasion Techniques}\label{sec:bg}

\begin{figure}[h]
    \centering
    \begin{tabular}{|l|}\hline
    \includegraphics[width=4in]{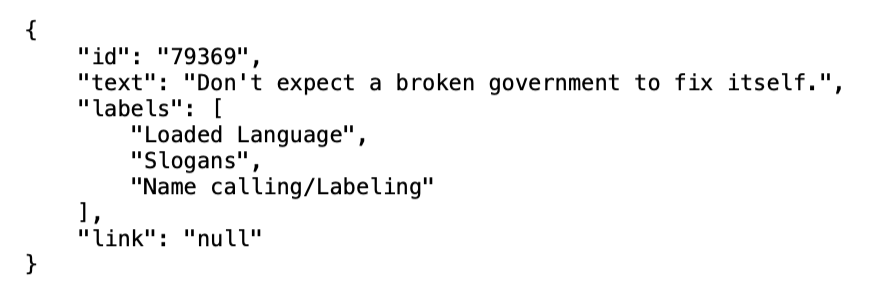} \\ \hline
    \end{tabular}
    \caption{A sample training instance. The text is labelled with three techniques, \textit{Loaded Language}, \textit{Slogans} and \textit{Name calling/Labelling}.}
    \label{fig:input}
\end{figure}

The goal of our model was to categorize the textual content of memes into one or several persuasion techniques. For example, given the training instance shown in Figure~\ref{fig:input}, the model needs to learn that the text \textit{Don't expect a broken government to fix itself} should be labelled with the three techniques provided in the labels field.

\begin{figure}[h]
    \centering
    \includegraphics[width=\columnwidth]{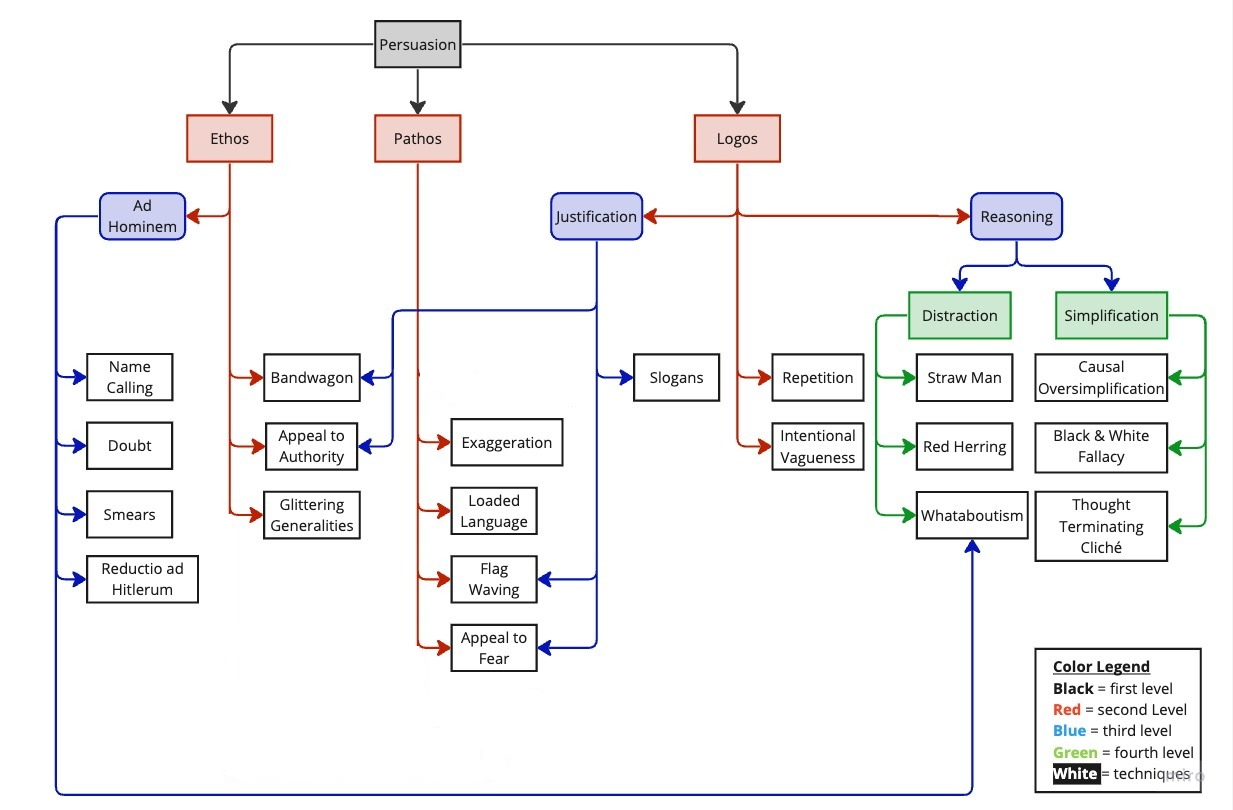}
    \caption{Hierarchy of the persuasion techniques to be used to label the texts. Figure slightly modified from~\cite{semeval2024task4}.}
    \label{fig:hierarchy}
\end{figure}

\paragraph{Persuasion Techniques:} 
The SemEval organizers provided an inventory of 20 persuasion techniques to be used as labels (eg: \textit{Loaded Language}, \textit{Slogans}, \textit{Name calling/Labelling}) and were structured hierarchically as shown in Figure~\ref{fig:hierarchy}. This rendered the task a hierarchical multi-label classification problem and was therefore evaluated using hierarchical precision, recall and F measures.

\begin{figure}[h]
    \centering
    \includegraphics[width=\columnwidth]{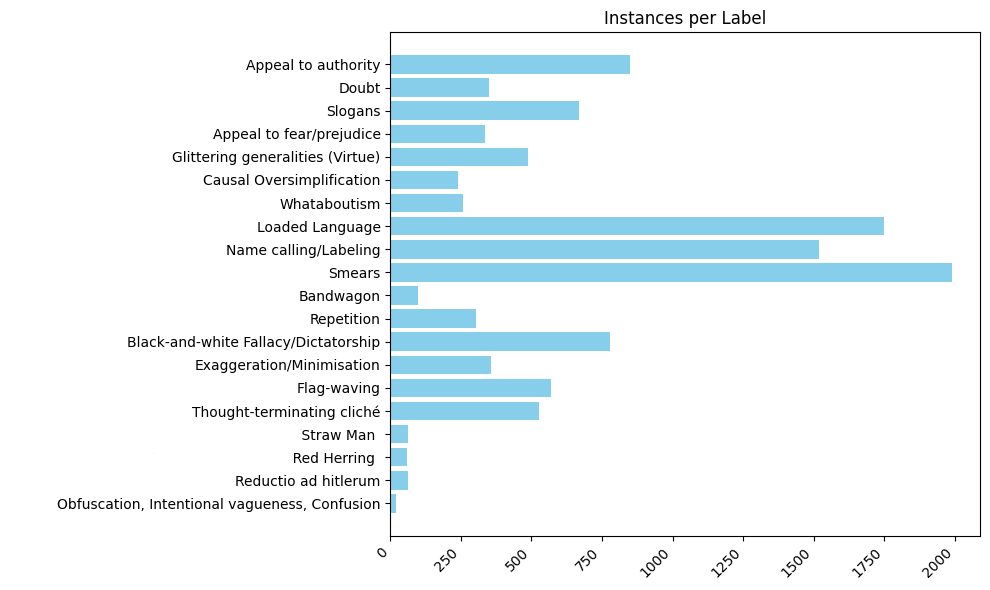}
    \caption{Distribution of the data for each persuasion technique in the training set.}
    \label{fig:original}
\end{figure}

\paragraph{Datasets:} 
The dataset provided contained memes collected from online public groups discussing a variety of topics such as politics, vaccines, COVID-19, gender equality, and the Russo-Ukrainian War. For our task, only the text extracted from these memes was used. The training (7k samples), validation (500 samples) and development (1k samples) sets included only English texts.\\Figure~\ref{fig:original} shows the distribution of instances for each persuasion technique in the training set. As the figure shows, some techniques, such as \textit{Loaded Language} and \textit{Smears}, had large number of instances in the training set (1750 and 1990 respectively); while others like \textit{Straw Man} and \textit{Red Herring} were severely underrepresented (62 and 59 instances respectively).

\begin{figure}[h]
    \centering
    \includegraphics[width=3in]{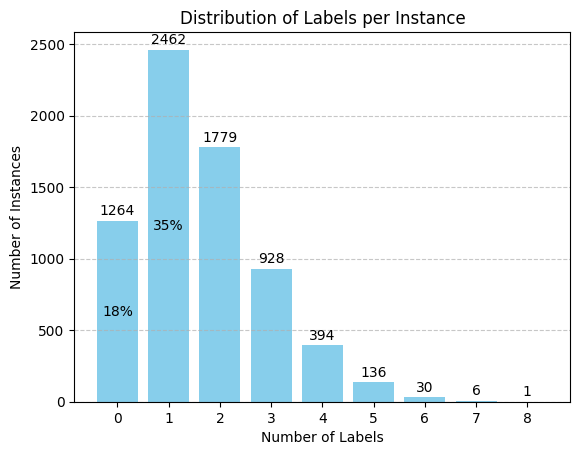}
    \caption{Distribution of persuasion techniques per instance in the training set.}
    \label{fig:distn}
\end{figure}

Figure~\ref{fig:distn} shows the distribution of labels per instances. As the figure shows, most of the instances (47\%) were labelled with multiple techniques, 35\% were labeled with only 1 technique and 18\% had no labels at all.\\Given the above English training set and hierarchical persuasion techniques, the goal of our model was to identify 0 or $n$ techniques for each textual instance in English and in 3 surprise languages.

\section{Proposed Approach}\label{sec:sys}
\begin{figure}[h]
    \centering
    \includegraphics[width=4in]{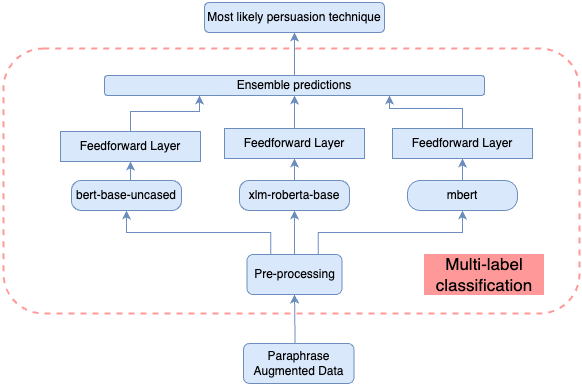}
    \caption{Schematic overview of our classification pipeline for the detection of persuasion techniques in memes.}
    \label{fig:model}
\end{figure}

Figure~\ref{fig:model} shows an overview of the classification pipeline we employed for this task. As shown in Figure~\ref{fig:model}, our methodology is based on fine-tuning three distinct pre-trained language models: BERT~\cite{devlin-etal-2019-bert}, XLM-RoBERTa~\cite{conneau-etal-2020-unsupervised}, and mBERT~\cite{devlin-etal-2019-bert} on augmented datasets.

\subsection{Multi-label Classification}\label{sec:multilabel}
As Figure~\ref{fig:model} shows, the data is first preprocessed  using standard tokenization. Then we proceeded to fine-tune three distinct models: \texttt{bert-base-uncased}, \texttt{xlm-roberta-base}, and \texttt{bert-base-multilingual-uncased} which returned a probability distribution over the 20 techniques. These three model predictions were then pooled via averaging.

Despite the hierarchical organization of the persuasion techniques, we opted to predicting solely the technique names (leaf nodes in Figure~\ref{fig:hierarchy}) and not their ancestor nodes. However, to address the multi-label classification, we implemented thresholding in order to determine which techniques have a high enough score to be part of the output label set. We experimented with custom values for each technique with values ranging from 0.01 to 0.7 and picked the optimal values for each class based on the validation set. These thresholds were applied to the scores obtained after passing the logits of each class through a sigmoid function.

To handle the three surprise languages, during the official testing phase system, the model trained only on English, would automatically translate the surprise language to English for our model's zero-shot predictions. This was inspired by the approach of~\cite{costa-etal-2023-clac}.

\begin{figure}[h]
    \centering
    \includegraphics[width=5.5in]{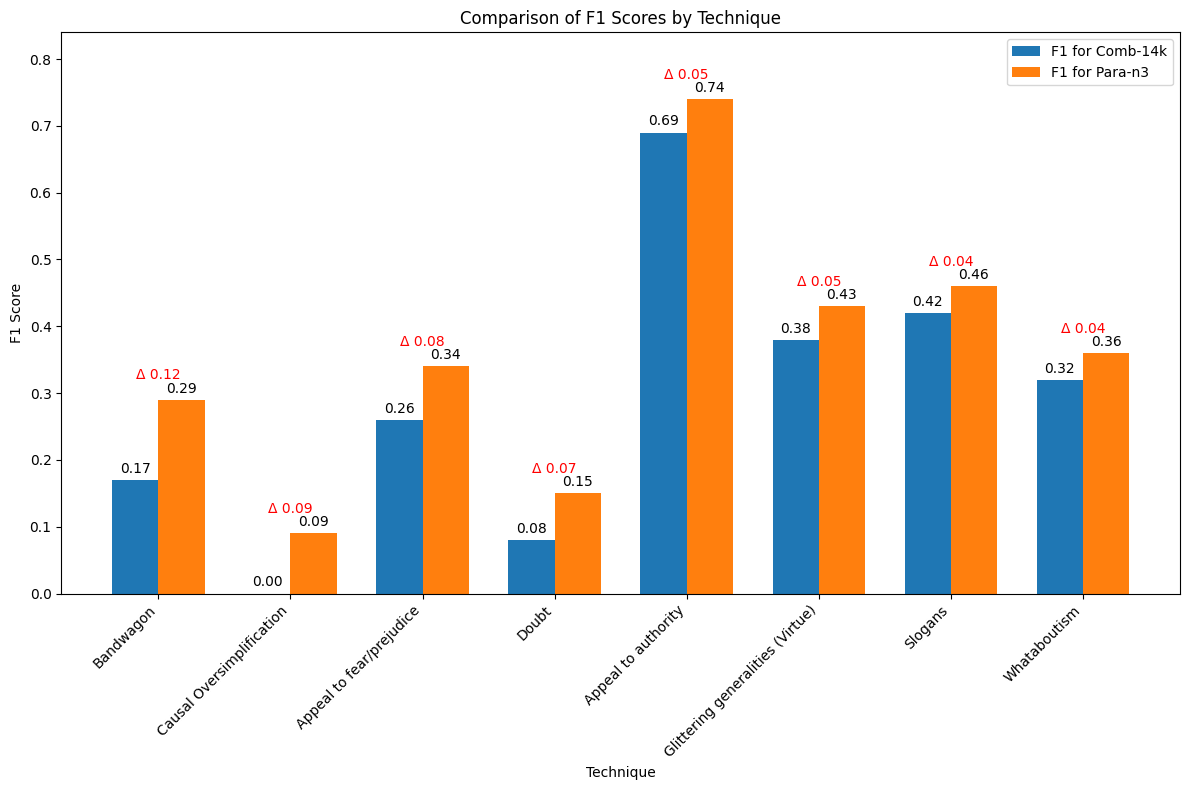}
    \caption{Techniques that showed an improvement in F1 score with the validation set when using $n$=3 paraphrases (i.e. \texttt{Para-n3}) compared to \texttt{Comb-14k}.}
    \label{fig:benefit-pic}
\end{figure}

\subsection{Data Augmentation}
\label{sec:data}
 To mitigate the lack of data we took advantage of two data augmentation strategies: an external dataset and automatically generated paraphrases.

\subsubsection{External Dataset (\texttt{Comb-14k} dataset):} \label{Sem2020}
\begin{figure}[h]
\begin{tabular}{cc}
    \hspace*{-1.5cm}\includegraphics[height=1.8in]{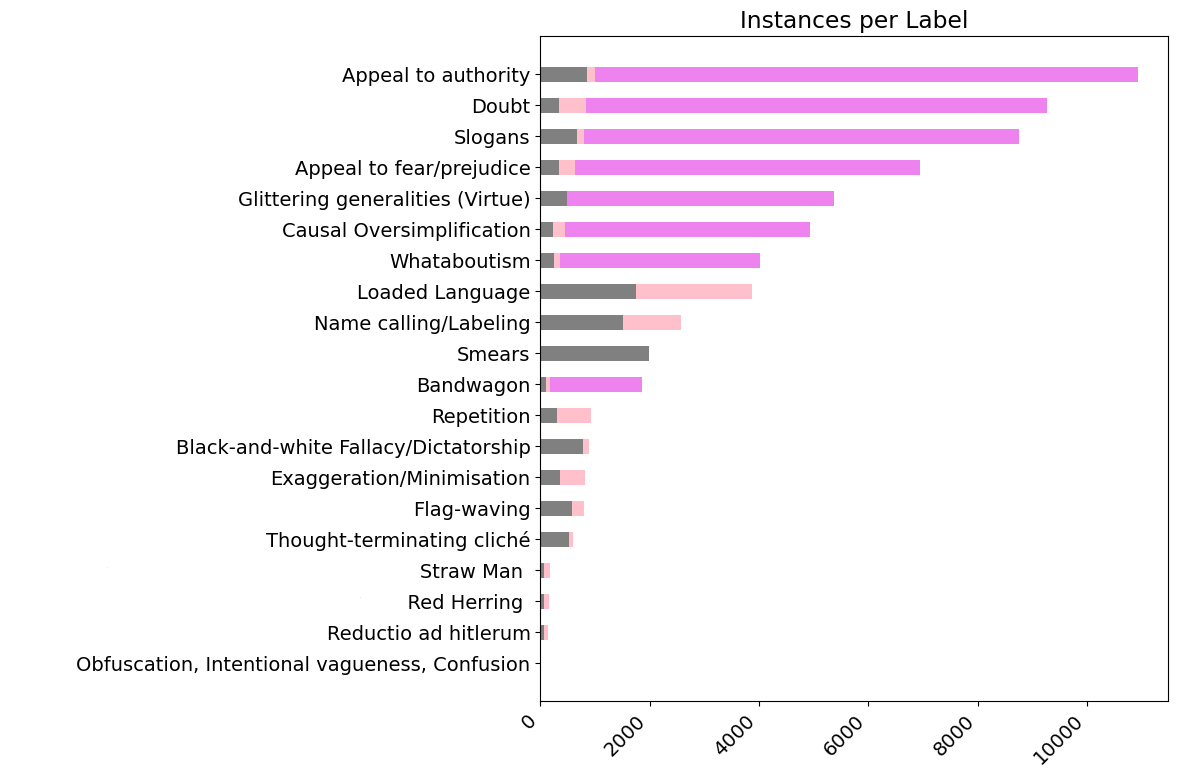}&\includegraphics[height=1.8in]{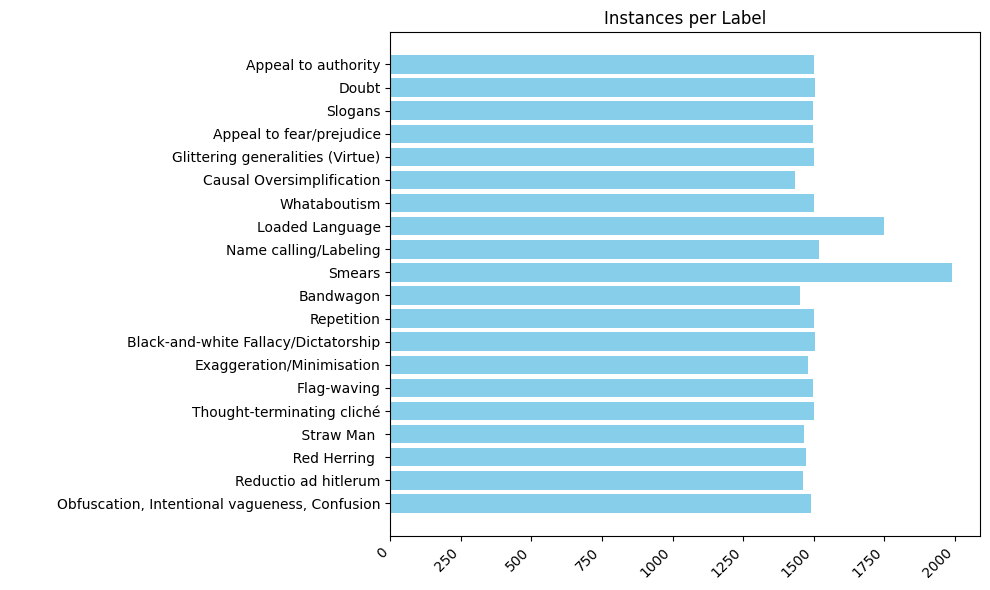 }\\
    (a) & (b) \\
    \end{tabular}
    \caption{Distribution of the data for each persuasion technique in (a) the original training set (in gray), the \texttt{Comb-14k} dataset (in gray + pink) and the \texttt{Para-Benef} dataset (in gray + pink + violet) (b) the \texttt{Para-Bal} dataset.}
    \label{fig:data_g3}
\end{figure}

The Technique Classification (TC) subtask from the SemEval~2020 Task~11~\cite{da-san-martino-etal-2020-semeval} provided a dataset of 7k instances from the news domain annotated with similar guidelines as this year's. In contrast to the 2020 task, this year's dataset covered a different domain and used a revised set of persuasion techniques compared to the 2020 inventory. Indeed, in the 2020 TC dataset, a few techniques were merged into a single category due to lack of data, resulting in a list of 14 techniques. In the current year, an expanded inventory of 20 techniques was employed. To ensure consistency between the two sets, we preprocessed the 2020 TC dataset by splitting techniques that had previously been merged. For example, we singled out \textit{Bandwagon} and \textit{Reductio ad Hitlerum}, which had been merged into a single technique in the SemEval 2020 TC dataset.\\We considered two approaches to leverage the modified 2020 TC dataset. The initial option involved pre-training models on this dataset, followed by fine-tuning on the 2024 training data—an approach implemented by~\cite{tian-etal-2021-mind}. Another approach entailed combining both datasets and fine-tuning models on this combined dataset. We chose the latter method because the two datasets covered different genres and a joint training approach would likely enable the model to better adapt and grasp nuanced linguistic patterns across both. For easy reference in the rest of the paper, we call the combined dataset \texttt{Comb-14k}. Figure~\ref{fig:data_g3}(a) (gray + pink) shows the resulting distribution of the persuasion techniques in this combined dataset.

\subsubsection{Paraphrasing (\texttt{Para-*} datasets)}
\label{para}
Despite having almost doubled each class with the use of the 2020 TC dataset, some classes were still severely underrepresented; see Figure~\ref{fig:data_g3}(a) (gray + pink). To address this, we took advantage of an LLM to generate paraphrases for each training instance, then labeled these paraphrases with the same set of labels as the original instance. Our intuition was twofold. First, generating paraphrases would expose the model to a more extensive set of samples for each class, potentially improving its ability to discern subtle nuances within the data. Second, paraphrasing sentences could unveil hidden semantics, providing the model with a tool to identify propaganda techniques reliant on nuanced linguistic choices or phrasing.
To generate paraphrases, we leveraged ChatGPT-3.5 turbo, setting the temperature  to 0.7. This value aimed to introduce diversity in the paraphrases while maintaining relevance to the original instances.
Several datasets were created using this method:

\begin{description}
\item[\texttt{Para-n1} and \texttt{Para-n3}:]
For each instance in \texttt{Comb-14k}, we generated $n$ paraphrases. We experimented with $n$=1 and $n$=3 leading to datasets of 28k and 52k respectively, which we call \texttt{Para-n1} and \texttt{Para-n3} respectively. \\The overall hierarchical F-score with the validation set given (500 instances) showed an increase when training with these datasets and $n=3$ seemed to perform better than $n=1$.\\However, a per-class analysis showed that not all classes benefited from the increase in support. For example, the persuasion technique \textit{Bandwagon} increased its F1 from 0.17 to 0.29; whereas \textit{Repetition} decreased its F1 from 0.56 to 0.31. We therefore identified the classes with improvement in F-score greater than 0.03 when using the \texttt{Para-n3} dataset compared to the \texttt{Comb-14k} dataset. These 8 techniques along with their increase in F-scores are shown in Figure~\ref{fig:benefit-pic}. This set of techniques formed the basis for our subsequent strategy.

\item[\texttt{Para-Benef}:]
Since only 8 techniques seemed to benefit from the use of paraphrases, we created a new augmented dataset by increasing the number of paraphrases only for these techniques. Specifically, let  $\mathbf{B}$ be the set of 8 techniques that benefited from paraphrases (see Figure~\ref{fig:benefit-pic}), for all data instances $d$ in \texttt{Comb-14k} labeled with techniques $\mathbf{T}=\{t_1, t_2,\ldots t_n\}$ (where $n\le20$), for each $t_i \in$ $\mathbf{B}$, we generated 10 paraphrases of $d$ and labeled them with all techniques from $\mathbf{T} \cap \mathbf{B}$.
This newly created dataset called \texttt{Para-Benef}, contained 54k instances.\\
Figure~\Ref{fig:data_g3}(a) shows the distribution of instances for each technique in the \texttt{Para-Benef} dataset (gray + pink + violet), in comparison with the original training set and the \texttt{Comb-14k} dataset. As Figure~\ref{fig:data_g3} shows, all datasets are severely imbalanced. Our next dataset therefore tried to address this issue.

\item[\texttt{Para-Bal}:]
Our last dataset used our paraphrase generation strategy to address the dataset imbalance. We rectified the underrepresented classes in the initial training dataset by augmenting them with paraphrases. The most frequent three techniques—\textit{Smears}, \textit{Name-calling/Labelling}, and \textit{Loaded Language}
had 1990, 1750, 1518 samples respectively. We thus aimed at reaching similar number of instances for the other techniques. We balanced the dataset by generating batches of 5 paraphrases for each other technique to reach around 1500 instances. This newly created dataset called \texttt{Para-Bal} contained 49k instances (see Figure~\ref{fig:data_g3}(b)).

\end{description}

Table~\ref{tab:dev} shows the results of the validation with the optimal threshold for each class using the official SemEval scorer~\cite{semeval2024task4}. As Table~\ref{tab:dev} shows, the use of the original dataset (7k instances) achieved  a hierarchical F1 of at most 0.49 on the development set; whereas all augmented training sets led to higher performances. The best model with both the validation and the development set was the ensemble trained on the \texttt{Para-Bal} dataset which reached an hierarchical F1 of 0.59 and 0.61 respectively. Surprisingly, \texttt{Para-Benef} which contained 10 paraphrases for the benefited techniques did not perform better than using only 3 paraphrases for all techniques (\texttt{Para-n3}). This suggests that the excessive inclusion of paraphrases from a different distribution (memes versus news) may have led to too much noise in the data.

\begin{table*}[h]
\caption{Hierarchical F1 scores of our models, when trained on different English-language datasets for both the validation and development sets.}
\label{tab:dev}
\centering
\scalebox{.92}{
\begin{tabular}{|ll|r|r|}
\hline
\multicolumn{1}{|l|}{\textbf{Training Set}} &
  \textbf{Models} &
  \multicolumn{1}{l|}{\textbf{Validation Set}} &
  \multicolumn{1}{l|}{\textbf{Development Set}} \\ \hline
  \multicolumn{1}{|l|}{\multirow{4}{*}{\begin{tabular}[c]{@{}l@{}} \\ \texttt{Original} \\ (7k) \end{tabular}}} &
  BERT &
  0.42&
  0.43 \\ \cline{2-4} 
\multicolumn{1}{|l|}{} & XLM-RoBERTa    & 0.48 & 0.49 \\ \cline{2-4} 
\multicolumn{1}{|l|}{} & mBERT          & 0.48 & 0.48 \\ \cline{2-4} 
\multicolumn{1}{|l|}{} & Ensemble Model & 0.45 & 0.46 \\ \hline \hline
  
  \multicolumn{1}{|l|}{\multirow{4}{*}{\begin{tabular}[c]{@{}l@{}} \\ \texttt{Comb-14k} \\ (14k) \end{tabular}}} &
  BERT &
  0.52 &
  0.55 \\ \cline{2-4} 
\multicolumn{1}{|l|}{} & XLM-RoBERTa    & 0.53 & 0.54 \\ \cline{2-4} 
\multicolumn{1}{|l|}{} & mBERT          & 0.53 & 0.54 \\ \cline{2-4} 
\multicolumn{1}{|l|}{} & Ensemble Model & 0.53 & 0.56 \\ \hline \hline
\multicolumn{1}{|l|}{\multirow{4}{*}{\begin{tabular}[c]{@{}l@{}}\texttt{Para-n1} \\ (28k) \end{tabular}}} &
  BERT &
  0.55 &
  0.57 \\ \cline{2-4} 
\multicolumn{1}{|l|}{} & XLM-RoBERTa    & 0.57 & 0.54 \\ \cline{2-4} 
\multicolumn{1}{|l|}{} & mBERT          & 0.50  & 0.53 \\ \cline{2-4} 
\multicolumn{1}{|l|}{} & Ensemble Model & 0.55 & 0.56 \\ \hline \hline
\multicolumn{1}{|l|}{\multirow{4}{*}{\begin{tabular}[c]{@{}l@{}}\texttt{Para-n3} \\ (52k)\end{tabular}}} &
  BERT &
  0.54 &
  0.55 \\ \cline{2-4} 
\multicolumn{1}{|l|}{} & XLM-RoBERTa    & 0.54 & 0.54 \\ \cline{2-4} 
\multicolumn{1}{|l|}{} & mBERT          & 0.54 & 0.55 \\ \cline{2-4} 
\multicolumn{1}{|l|}{} &
  Ensemble Model &
0.56 &
0.57 \\ \hline \hline
\multicolumn{1}{|l|}{\multirow{4}{*}{\begin{tabular}[c]{@{}l@{}}\texttt{Para-Benef}\\ (54k) \end{tabular}}} &
  BERT &
  0.48 &
  0.51 \\ \cline{2-4} 
\multicolumn{1}{|l|}{} & XLM-RoBERTa    & 0.54 & 0.55 \\ \cline{2-4} 
\multicolumn{1}{|l|}{} & mBERT          & 0.51 & 0.53 \\ \cline{2-4} 
\multicolumn{1}{|l|}{} & Ensemble Model & 0.54 & 0.55 \\ \hline \hline
\multicolumn{1}{|l|}{\multirow{4}{*}{\begin{tabular}[c]{@{}l@{}}\texttt{Para-Bal}\\ (49k) \end{tabular}}} &
  BERT &
  0.54 &
  0.58 \\ \cline{2-4} 
\multicolumn{1}{|l|}{} & XLM-RoBERTa    & 0.58 & 0.59 \\ \cline{2-4} 
\multicolumn{1}{|l|}{} & mBERT          & 0.53 & 0.55 \\ \cline{2-4} 
\multicolumn{1}{|l|}{} & Ensemble Model & \textbf{0.59} & \textbf{0.61}\\ \hline \hline
\end{tabular}%
}
\end{table*}

\begin{figure}[h]
    \centering
    \includegraphics[width=5.5in]{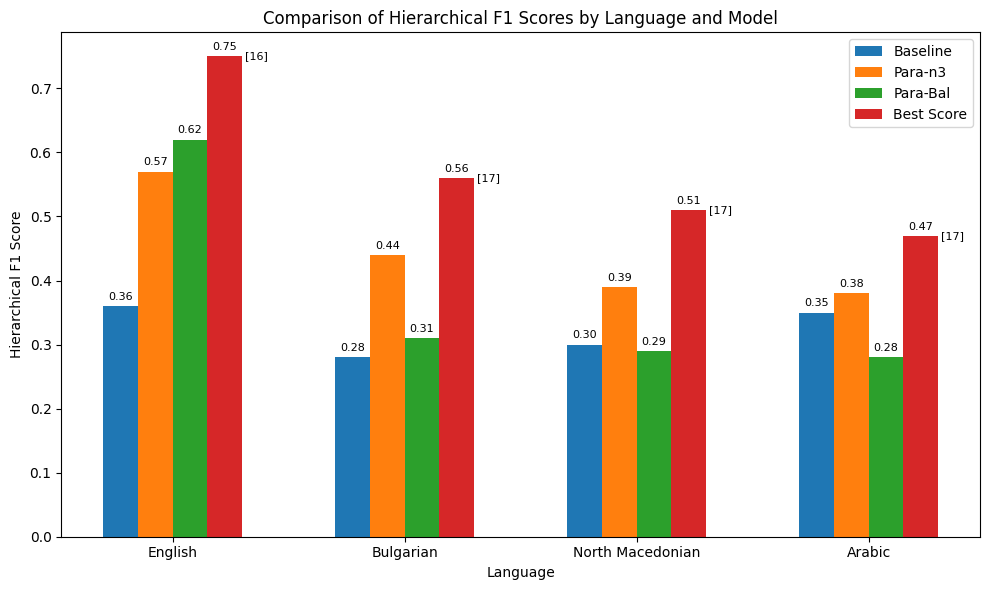}
    \caption{Comparison of the final hierarchical F1 scores obtained by our official model
    trained with \texttt{Para-n3}, the model trained with \texttt{Para-Bal}, the best corresponding models (\cite{914isthebest2024task4} for English, \cite{OtterlyObsessedWithSemanticsSemeval2024task4} for Bulgarian, North Macedonian and Arabic) in the shared task and the baseline in each given language.}
    \label{fig:test-comp}
\end{figure}

\section{Experimental Setup} \label{sec:exp}

\subsection{System Pipeline and Training Details} \label{subsec: model-info}
The system pipeline code was implemented in PyTorch. The pre-trained models BERT [\texttt{bert-base-uncased}], XLM-RoBERTa [\texttt{xlm-roberta-base}], and mBERT [\texttt{bert-base-multilingual-uncased}] and their tokenizers were sourced from Hugging Face. All models were trained for 10 epochs using the Adam optimizer with a learning rate of 2e-5. Batch sizes varied with BERT utilizing 128, and XLM-RoBERTa and mBERT using 64. A final feedforward layer with 20 logits (equal to the number of persuasion techniques) was added to each model. The Binary Cross Entropy with logits served as the loss function, with one-hot encoding applied to the true labels. For prediction, a sigmoid activation function was used on the logits, followed by thresholding. The ensemble model used an unweighted average of all predictions from the three individual models. The
ChatGPT-3.5 turbo\footnote{\url{https://platform.openai.com/docs/models/gpt-3-5-turbo}} API with a temperature set to 0.7 was used for paraphrase generation. During testing, the surprise languages were translated into English using the deep-translator API\footnote{\url{https://pypi.org/project/deep-translator/}}.

\section{Results and Analysis}\label{sec:res}

For our official submission to SemEval, the \texttt{Para-Bal} dataset had not been created yet; hence our official results are based on the ensemble model trained on the union of \texttt{Para-n3} and the development set (1k samples), for a total of 53k samples. The three surprise languages were Bulgarian, North Macedonian and Arabic. The test set contained 1500 samples for English, 426 samples for Bulgarian, 259 samples for North Macedonian and 100 samples for Arabic.
The official results of our system are shown in Figure~\ref{fig:test-comp}, along with a baseline score that assigns the most frequent persuasion technique to all instances, and the score obtained by the best performing systems for each language~\cite{914isthebest2024task4,OtterlyObsessedWithSemanticsSemeval2024task4}. As Figure~\ref{fig:test-comp} shows, although our ensemble model did not reach the top performance for English (0.57 versus 0.75), it performed better than the baseline in all languages except Arabic, where the improvement was not significant.

Using the same testing protocol, we reproduced the results using the model trained with the balanced training dataset (\texttt{Para-Bal}). The results displayed in Figure~\ref{fig:test-comp} indicate an improvement in score with the English test set (0.62 versus 0.57). This again confirms the importance of a balanced dataset, and paraphrases based on the same distribution as the original texts. Indeed, although \texttt{Para-n3} is larger than \texttt{Para-Bal} (52k versus 49k), it is not balanced and contains paraphrases of instances from different genres (memes and news). However, surprisingly, the performance enhancement when using \texttt{Para-Bal} is not observed for the zero-shot classification of the surprise languages whose performance dropped significantly. For these languages, a larger training set, even with noisy out-of-distribution instances, leads to better results possibly due to the noise introduced by the automatic translation itself.

Compared to the other approaches at the 2024 edition of SemEval Task~4, the top performing team overall, OtterlyObsessedWithSemantics~\cite{OtterlyObsessedWithSemanticsSemeval2024task4} designed a specialized classification head to enhance a Large Language Model. They organized the structure across several connected layers, enabling them to build upon earlier decisions in later, more detailed layers. They optimized the model's performance by systematically exploring different hyperparameters through grid-search. In addition, similarly to our approach, during the testing phase, they translated all the surprise language datasets into English. \cite{914isthebest2024task4} developed a system using Chain-of-Thought based data augmentation methods, in-domain pre-training and ensemble strategy that combined the strengths of both RoBERTa and DeBERTa models.

\section{Conclusion and Future Work}\label{sec:conc}

This paper described our approach to hierarchical multi-label detection of persuasion techniques in meme texts. We used an ensemble model with three fine-tuned language models and incorporated data augmentation through paraphrasing from ChatGPT. We tested our approach through the SemEval 2024 Task~4 subtask~1~\cite{semeval2024task4}. During testing, our system outperformed the baseline in all languages. Analysis of the results show the importance of dataset balancing and paraphrasing techniques in enhancing model performance. Despite having a smaller number of instances, the balanced dataset consistently outperforms its unbalanced counterparts, demonstrating the efficacy of balancing methods. Moreover, data augmentation improves model performance, as indicated by the under-performance of models trained on the original dataset (7k instances). Additionally, the results underscore the potential drawbacks of including paraphrases from diverse distributions, which may introduce significant noise into the system, potentially compromising overall effectiveness. This study prompts further inquiry into the specific drivers of performance improvement, whether it be dataset balancing or the inclusion of external data. Although our zero-shot approach exhibits limitations, it underscores the positive correlation between data volume and model performance, as illustrated by the superior performance of models trained on larger paraphrased datasets, such as \texttt{Para-n3} with 56k instances compared to \texttt{Para-Bal} with 49k instances

Moving forward, to boost performance, it would be interesting to measure the influence of the quality and similarity of the paraphrases on the performance. Moreover, incorporating hierarchical predictions, possibly at a second-level node, could improve scores further. Exploring the utilization of larger multilingual models alongside language-specific datasets and experimenting with various ensemble methods could be fruitful. Finally, considering the integration of adversarial training or self-supervised learning techniques might offer valuable avenues for improvement.

\section*{Acknowledgements}
The authors would like to thank the organisers of the SemEval shared task. This work was financially supported by the Natural Sciences and Engineering Research Council of Canada (NSERC).

\bibliographystyle{IEEEtran}
\bibliography{references}

\begin{thebibliography}{10}
\providecommand{\url}[1]{#1}
\csname url@samestyle\endcsname
\providecommand{\newblock}{\relax}
\providecommand{\bibinfo}[2]{#2}
\providecommand{\BIBentrySTDinterwordspacing}{\spaceskip=0pt\relax}
\providecommand{\BIBentryALTinterwordstretchfactor}{4}
\providecommand{\BIBentryALTinterwordspacing}{\spaceskip=\fontdimen2\font plus
\BIBentryALTinterwordstretchfactor\fontdimen3\font minus \fontdimen4\font\relax}
\providecommand{\BIBforeignlanguage}[2]{{%
\expandafter\ifx\csname l@#1\endcsname\relax
\typeout{** WARNING: IEEEtran.bst: No hyphenation pattern has been}%
\typeout{** loaded for the language `#1'. Using the pattern for}%
\typeout{** the default language instead.}%
\else
\language=\csname l@#1\endcsname
\fi
#2}}
\providecommand{\BIBdecl}{\relax}
\BIBdecl

\bibitem{semeval2024task4}
D.~Dimitrov, F.~Alam, M.~Hasanain, A.~Hasnat, F.~Silvestri, P.~Nakov, and G.~Da~San~Martino, ``Semeval-2024 task 4: Multilingual detection of persuasion techniques in memes,'' in \emph{Proceedings of the 18th International Workshop on Semantic Evaluation}, ser. SemEval 2024, Mexico City, Mexico, June 2024.

\bibitem{jurkiewicz-etal-2020-applicaai}
\BIBentryALTinterwordspacing
D.~Jurkiewicz, {\L}.~Borchmann, I.~Kosmala, and F.~Grali{\'n}ski, ``{A}pplica{AI} at {S}em{E}val-2020 task 11: On {R}o{BERT}a-{CRF}, span {CLS} and whether self-training helps them,'' in \emph{Proceedings of the Fourteenth Workshop on Semantic Evaluation}, A.~Herbelot, X.~Zhu, A.~Palmer, N.~Schneider, J.~May, and E.~Shutova, Eds.\hskip 1em plus 0.5em minus 0.4em\relax Barcelona (online): International Committee for Computational Linguistics, Dec. 2020, pp. 1415--1424. [Online]. Available: \url{https://aclanthology.org/2020.semeval-1.187}
\BIBentrySTDinterwordspacing

\bibitem{tian-etal-2021-mind}
\BIBentryALTinterwordspacing
J.~Tian, M.~Gui, C.~Li, M.~Yan, and W.~Xiao, ``{M}in{D} at {S}em{E}val-2021 task 6: Propaganda detection using transfer learning and multimodal fusion,'' in \emph{Proceedings of the 15th International Workshop on Semantic Evaluation (SemEval-2021)}, A.~Palmer, N.~Schneider, N.~Schluter, G.~Emerson, A.~Herbelot, and X.~Zhu, Eds.\hskip 1em plus 0.5em minus 0.4em\relax Online: Association for Computational Linguistics, Aug. 2021, pp. 1082--1087. [Online]. Available: \url{https://aclanthology.org/2021.semeval-1.150}
\BIBentrySTDinterwordspacing

\bibitem{chalkidis-etal-2019-large}
\BIBentryALTinterwordspacing
I.~Chalkidis, E.~Fergadiotis, P.~Malakasiotis, and I.~Androutsopoulos, ``Large-scale multi-label text classification on {EU} legislation,'' in \emph{Proceedings of the 57th Annual Meeting of the Association for Computational Linguistics}, A.~Korhonen, D.~Traum, and L.~M{\`a}rquez, Eds.\hskip 1em plus 0.5em minus 0.4em\relax Florence, Italy: Association for Computational Linguistics, Jul. 2019, pp. 6314--6322. [Online]. Available: \url{https://aclanthology.org/P19-1636}
\BIBentrySTDinterwordspacing

\bibitem{lin-etal-2023-effective}
\BIBentryALTinterwordspacing
N.~Lin, G.~Qin, G.~Wang, D.~Zhou, and A.~Yang, ``An effective deployment of contrastive learning in multi-label text classification,'' in \emph{Findings of the Association for Computational Linguistics: ACL 2023}, A.~Rogers, J.~Boyd-Graber, and N.~Okazaki, Eds.\hskip 1em plus 0.5em minus 0.4em\relax Toronto, Canada: Association for Computational Linguistics, Jul. 2023, pp. 8730--8744. [Online]. Available: \url{https://aclanthology.org/2023.findings-acl.556}
\BIBentrySTDinterwordspacing

\bibitem{SemEval2018Task1}
S.~M. Mohammad, F.~Bravo-Marquez, M.~Salameh, and S.~Kiritchenko, ``Semeval-2018 {T}ask 1: {A}ffect in tweets,'' in \emph{Proceedings of International Workshop on Semantic Evaluation (SemEval-2018)}, New Orleans, LA, USA, 2018.

\bibitem{alhuzali-ananiadou-2021-spanemo}
\BIBentryALTinterwordspacing
H.~Alhuzali and S.~Ananiadou, ``{S}pan{E}mo: Casting multi-label emotion classification as span-prediction,'' in \emph{Proceedings of the 16th Conference of the European Chapter of the Association for Computational Linguistics: Main Volume}, P.~Merlo, J.~Tiedemann, and R.~Tsarfaty, Eds.\hskip 1em plus 0.5em minus 0.4em\relax Online: Association for Computational Linguistics, Apr. 2021, pp. 1573--1584. [Online]. Available: \url{https://aclanthology.org/2021.eacl-main.135}
\BIBentrySTDinterwordspacing

\bibitem{CERRI201439}
\BIBentryALTinterwordspacing
R.~Cerri, R.~C. Barros, and A.~C. {de Carvalho}, ``Hierarchical multi-label classification using local neural networks,'' \emph{Journal of Computer and System Sciences}, vol.~80, no.~1, pp. 39--56, 2014. [Online]. Available: \url{https://www.sciencedirect.com/science/article/pii/S0022000013000718}
\BIBentrySTDinterwordspacing

\bibitem{zhou-etal-2020-hierarchy}
\BIBentryALTinterwordspacing
J.~Zhou, C.~Ma, D.~Long, G.~Xu, N.~Ding, H.~Zhang, P.~Xie, and G.~Liu, ``Hierarchy-aware global model for hierarchical text classification,'' in \emph{Proceedings of the 58th Annual Meeting of the Association for Computational Linguistics}, D.~Jurafsky, J.~Chai, N.~Schluter, and J.~Tetreault, Eds.\hskip 1em plus 0.5em minus 0.4em\relax Online: Association for Computational Linguistics, Jul. 2020, pp. 1106--1117. [Online]. Available: \url{https://aclanthology.org/2020.acl-main.104}
\BIBentrySTDinterwordspacing

\bibitem{Yu2022ConstrainedSG}
\BIBentryALTinterwordspacing
C.~Yu, Y.~Shen, Y.~Mao, and L.~Cai, ``Constrained sequence-to-tree generation for hierarchical text classification,'' \emph{Proceedings of the 45th International ACM SIGIR Conference on Research and Development in Information Retrieval}, 2022. [Online]. Available: \url{https://api.semanticscholar.org/CorpusID:247939715}
\BIBentrySTDinterwordspacing

\bibitem{da-san-martino-etal-2020-semeval}
\BIBentryALTinterwordspacing
G.~Da~San~Martino, A.~Barr{\'o}n-Cede{\~n}o, H.~Wachsmuth, R.~Petrov, and P.~Nakov, ``{{S}em{E}val-2020 Task 11: Detection of Propaganda Techniques in News Articles},'' in \emph{Proceedings of the Fourteenth Workshop on Semantic Evaluation}.\hskip 1em plus 0.5em minus 0.4em\relax Barcelona (online): International Committee for Computational Linguistics, Dec. 2020, pp. 1377--1414. [Online]. Available: \url{https://aclanthology.org/2020.semeval-1.186}
\BIBentrySTDinterwordspacing

\bibitem{dimitrov-etal-2021-semeval}
\BIBentryALTinterwordspacing
D.~Dimitrov, B.~Bin~Ali, S.~Shaar, F.~Alam, F.~Silvestri, H.~Firooz, P.~Nakov, and G.~Da~San~Martino, ``{S}em{E}val-2021 task 6: Detection of persuasion techniques in texts and images,'' in \emph{Proceedings of the 15th International Workshop on Semantic Evaluation (SemEval-2021)}, A.~Palmer, N.~Schneider, N.~Schluter, G.~Emerson, A.~Herbelot, and X.~Zhu, Eds.\hskip 1em plus 0.5em minus 0.4em\relax Online: Association for Computational Linguistics, Aug. 2021, pp. 70--98. [Online]. Available: \url{https://aclanthology.org/2021.semeval-1.7}
\BIBentrySTDinterwordspacing

\bibitem{devlin-etal-2019-bert}
\BIBentryALTinterwordspacing
J.~Devlin, M.-W. Chang, K.~Lee, and K.~Toutanova, ``{{BERT}: Pre-training of Deep Bidirectional Transformers for Language Understanding},'' in \emph{Proceedings of the 2019 Conference of the North {A}merican Chapter of the Association for Computational Linguistics: Human Language Technologies, Volume 1 (Long and Short Papers)}.\hskip 1em plus 0.5em minus 0.4em\relax Minneapolis, Minnesota, USA: Association for Computational Linguistics, Jun. 2019, pp. 4171--4186. [Online]. Available: \url{https://aclanthology.org/N19-1423}
\BIBentrySTDinterwordspacing

\bibitem{conneau-etal-2020-unsupervised}
\BIBentryALTinterwordspacing
A.~Conneau, K.~Khandelwal, N.~Goyal, V.~Chaudhary, G.~Wenzek, F.~Guzm{\'a}n, E.~Grave, M.~Ott, L.~Zettlemoyer, and V.~Stoyanov, ``{Unsupervised Cross-lingual Representation Learning at Scale},'' in \emph{Proceedings of the 58th Annual Meeting of the Association for Computational Linguistics}.\hskip 1em plus 0.5em minus 0.4em\relax Online: Association for Computational Linguistics, Jul. 2020, pp. 8440--8451. [Online]. Available: \url{https://aclanthology.org/2020.acl-main.747}
\BIBentrySTDinterwordspacing

\bibitem{costa-etal-2023-clac}
\BIBentryALTinterwordspacing
N.~F. Costa, B.~Hamilton, and L.~Kosseim, ``{CL}a{C} at {S}em{E}val-2023 task 3: Language potluck {R}o{BERT}a detects online persuasion techniques in a multilingual setup,'' in \emph{Proceedings of the 17th International Workshop on Semantic Evaluation (SemEval-2023)}, A.~K. Ojha, A.~S. Do{\u{g}}ru{\"o}z, G.~Da~San~Martino, H.~Tayyar~Madabushi, R.~Kumar, and E.~Sartori, Eds.\hskip 1em plus 0.5em minus 0.4em\relax Toronto, Canada: Association for Computational Linguistics, Jul. 2023, pp. 1613--1618. [Online]. Available: \url{https://aclanthology.org/2023.semeval-1.223}
\BIBentrySTDinterwordspacing

\bibitem{914isthebest2024task4}
D.~Li, C.~Wang, X.~Zou, J.~Wang, P.~Chen, J.~Wang, L.~Yang, and H.~Lin, ``914isthebest at semeval-2024 task 4: Cot-based data augmentation strategy for persuasion techniques detection,'' in \emph{Proceedings of the 18th International Workshop on Semantic Evaluation}, ser. SemEval 2024, Mexico City, Mexico, June 2024.

\bibitem{OtterlyObsessedWithSemanticsSemeval2024task4}
J.~Wunderle, J.~Schubert, A.~Cacciatore, A.~Zehe, J.~Pfister, and A.~Hotho, ``Otterlyobsessedwithsemantics at semeval-2024 task 4: Developing a hierarchical multi-label classification head for large language models,'' in \emph{Proceedings of the 18th International Workshop on Semantic Evaluation}, ser. SemEval 2024, Mexico City, Mexico, June 2024.

\end{thebibliography}

\section*{Authors}
\noindent {\bf Kota Shamanth Ramanath Nayak} received B.Tech from Manipal Institute of Technology, India. Currently, he is pursuing his Master's in Computer Science
from Concordia University, Canada. His research interests include Natural Language Processing, Discourse Analysis and Large Language Models.\\

\noindent {\bf Leila Kosseim} obtained her PhD from the University of Montreal in 1995 on the topic of Natural Language Generation. Currently she is a professor in the Computer Science \& Software Engineering (CSSE) Department at Concordia University in Montreal. Her research interests include Natural Language Processing, Artificial Intelligence and Text Mining.

\end{document}